\documentclass{article}
\usepackage{spconf,amsmath,graphicx}
\usepackage{subfigure,multirow,amssymb,latexsym}

\title{COMBINING THE SILHOUETTE AND SKELETON DATA FOR GAIT RECOGNITION}
\name{Likai Wang \qquad Ruize Han$^{\ast}$ \qquad Wei Feng \thanks{*Corresponding author} \thanks{This work was supported by NSFC under Grants 62072334, U1803264.}}

\address{College of Intelligence and Computing, Tianjin University, Tianjin, China}
\begin{document}
%
\maketitle

\begin{abstract}
Gait recognition, a long-distance biometric technology, has aroused intense interest recently. Currently, the two dominant gait recognition works are appearance-based and model-based, which extract features from silhouettes and skeletons, respectively. However, appearance-based methods are greatly affected by clothes-changing and carrying conditions, while model-based methods are limited by the accuracy of pose estimation. To tackle this challenge, a simple yet effective two-branch network is proposed in this paper, which contains a CNN-based branch taking silhouettes as input and a GCN-based branch taking skeletons as input. In addition, for better gait representation in the GCN-based branch, we present a fully connected graph convolution operator to integrate multi-scale graph convolutions and alleviate the dependence on natural joint connections. Also, we deploy a multi-dimension attention module named STC-Att to learn spatial, temporal and channel-wise attention simultaneously. The experimental results on CASIA-B and OUMVLP show that our method achieves state-of-the-art performance in various conditions.
\end{abstract}
\section{Introduction}
\label{sec:intro}

Gait recognition is a research topic for systematic study on human motion, which recently attracts increasing interest in computer vision. Due to the fact that gait can be recognized at a long-distance without the cooperation of subjects while other biometric characteristics cannot, gait recognition has considerable prospect for many practical applications, e.g., video surveillance, crime investigation and social security. 

Currently, the two dominant gait recognition works are appearance-based and model-based.
The first extracts features from silhouettes using CNNs~\cite{huang2021context} or transformer network~\cite{xu2020cross,li2019joint}, which either compresses all silhouettes into one gait energy image (GEI) \cite{wu2016comprehensive,zhang2019learning} or regards gait as silhouette sequences \cite{huang2021context,chao2019gaitset,su2020deep,lin2021gait,chai2022lagrange}. However, such methods that rely on the human body shape are extremely sensitive to carrying, clothing and view-changing since these variations alter the human appearance drastically. In contrast to the above, the second category takes raw skeleton data obtained by pose estimation algorithms~\cite{cao2017realtime} as input and performs feature extraction through CNNs~\cite{an2018improving,liao2020model} or GCNs~\cite{teepe2022towards,wang2022frame}. Although such methods can overcome the interference caused by occlusion and view-changing to a certain extent, the accuracy of pose estimation methods and the lack of information contained in skeleton data bring limitations to the improvement of gait recognition accuracy. 
Recently, the work in~\cite{cai2021hybrid} has attempted to combine these two methods, however, complex models are used with weak performance improvement. 

Motivated by the above analysis, we propose a simple yet effective two-branch network for gait recognition task to realize the reciprocity of appearance- and model-based methods, which can significantly promote the enhancement of recognition accuracy. 
Specifically, the appearance-based branch leverages stacked CNNs to extract discriminative representations from a sequence of gait silhouettes. The model-based branch deploys GCNs to learn multi-order robust and compact representations from human skeleton data.

In addition, considering that existing GCN-based methods have not been well studied and have relatively low accuracy compared to appearance-based ones, we further design two modules for representing better skeleton feature in the GCN-based branch. 
Specifically, previous GCN-based methods~\cite{wang2022multi,teepe2022towards} learn the movement patterns of human joint from a spatial-temporal graph with joints as nodes and bones as edges. 
However, it is indisputable that the feature extractor should not only aggregate the information of directly connected joints, but also extract multi-scale structural features and long-range dependencies on account of strong correlations between joints that are physically apart. 
To achieve this, existing methods~\cite{wang2022multi} apply \textit{multiple adjacency matrices with different orders} for various joint connections, which suffers from the balance problem between training difficulty and the effective fusion of important information from distant joints, i.e., the \textit{distant-joint connection problem}. To address this problem, we present a fully-connected graph structure without natural joint connections, to adaptively integrate multi-scale graph convolutions instead of the redundant adjacency matrixes for distant joints. 
Moreover, inspired by the effectiveness of the attention mechanism demonstrated in recent appearance-based works~\cite{li2019attentive,dou2022metagait}, we propose an attention module named STC-Att in the model-based branch.
Extensive experiments indicate that our model achieves state-of-the-art performance in various conditions. 

The main contributions of the proposed method are summarized as follows:
(1) We propose a new two-branch framework for gait recognition task, which combines appearance- and model-based features to obtain their complementary advantages. 
(2) In the model-based branch, we propose an effective fully connected graph convolution operator, which can adaptively capture the dependencies between all nodes without multi-scale convolutions. Also, for more discriminative representations from raw skeleton data, we propose a multi-dimension attention module named STC-Att to learn the spatial, temporal and channel-wise attention simultaneously.
(3) Experimental results demonstrate that our method achieves superior performance, especially on the cloth-changing condition, than the prior state-of-the-art methods. 

\vspace{-3pt}
\section{Proposed method}
\label{sec:method}
\vspace{-3pt}

\subsection{A Two-branch Framework}
\label{sec:Model}

In this paper, we design a two-branch network to combine the appearance- and model-based methods, making the silhouettes and skeletons to be complementary with each other.

\begin{figure}[h]
\centering
\includegraphics[width=0.4\textwidth]{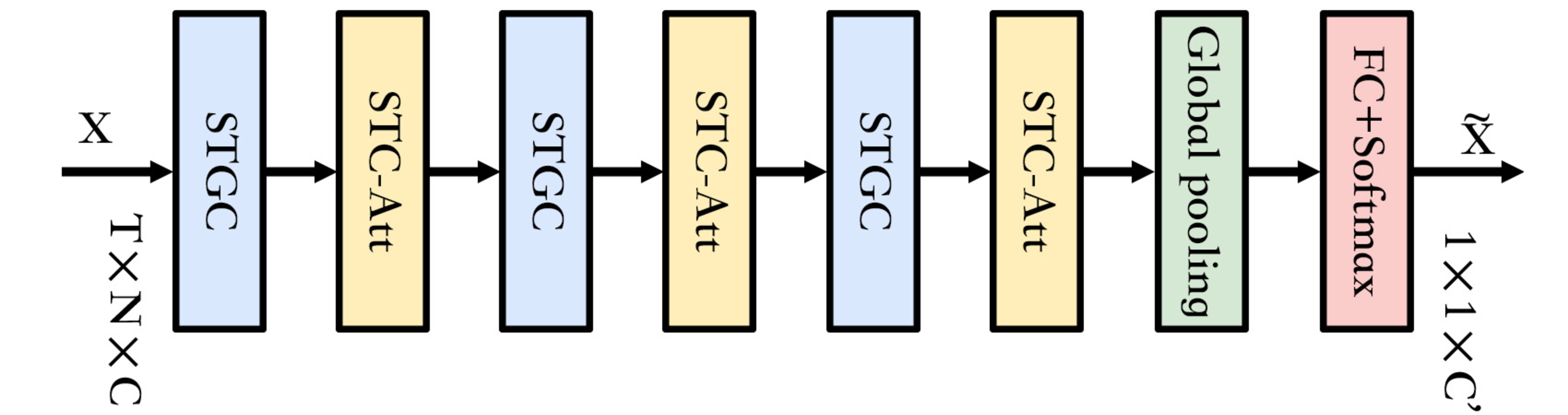}
\vspace{-11pt}
\caption{Architecture of the model-based branch.}
\label{fig:model}
\vspace{-5pt}
\end{figure}

The architecture of the model-based branch is shown in Figure~\ref{fig:model}.
As shown, we use the spatial-temporal graph convolutional (STGC) blocks~\cite{liu2020disentangling} as the basic component of the model.
In the STGC block, we apply the proposed fully-connected graph convolution (which will be discussed in Section~\ref{sec:Unified}) to capture the arbitrary-hop joint dependencies.
The proposed STC-Att module (which will be discussed in Section~\ref{sec:Attention}) is employed after each STGC block to explicitly model the channel-, spatial- and temporal-wise correlations between feature maps.
Finally, a global average pooling layer and a softmax classifier are utilized at the end of the model. 
In the appearance-based branch, we directly choose a state-of-the-art part-based model GaitPart~\cite{fan2020gaitpart} as the feature extractor. 

Suppose that the output feature vector of the model-based branch is $\textbf{f}_\mathrm{m}$ and of the appearance-based branch is $\textbf{f}_\mathrm{a}$, a straightforward concatenation operation is adopted to merge the two feature vectors together, which brings out the final gait descriptors for recognition. Note that because the two vectors, that are obtained from completely different raw data using diverse networks, have unequal vector size and value range, a ratio $\lambda$ is needed to adjust the rate of them. Hence, we can get the overall feature vector through $\textbf{f}=\textbf{f}_\mathrm{m} \oplus \lambda \cdot \textbf{f}_\mathrm{a}$, where $\oplus$ means the concatenation operation. 

\vspace{-3pt}
\subsection{Fully Connected Graph Convolution}
\label{sec:Unified}

\begin{figure}
\centering
\includegraphics[width=0.13\textwidth]{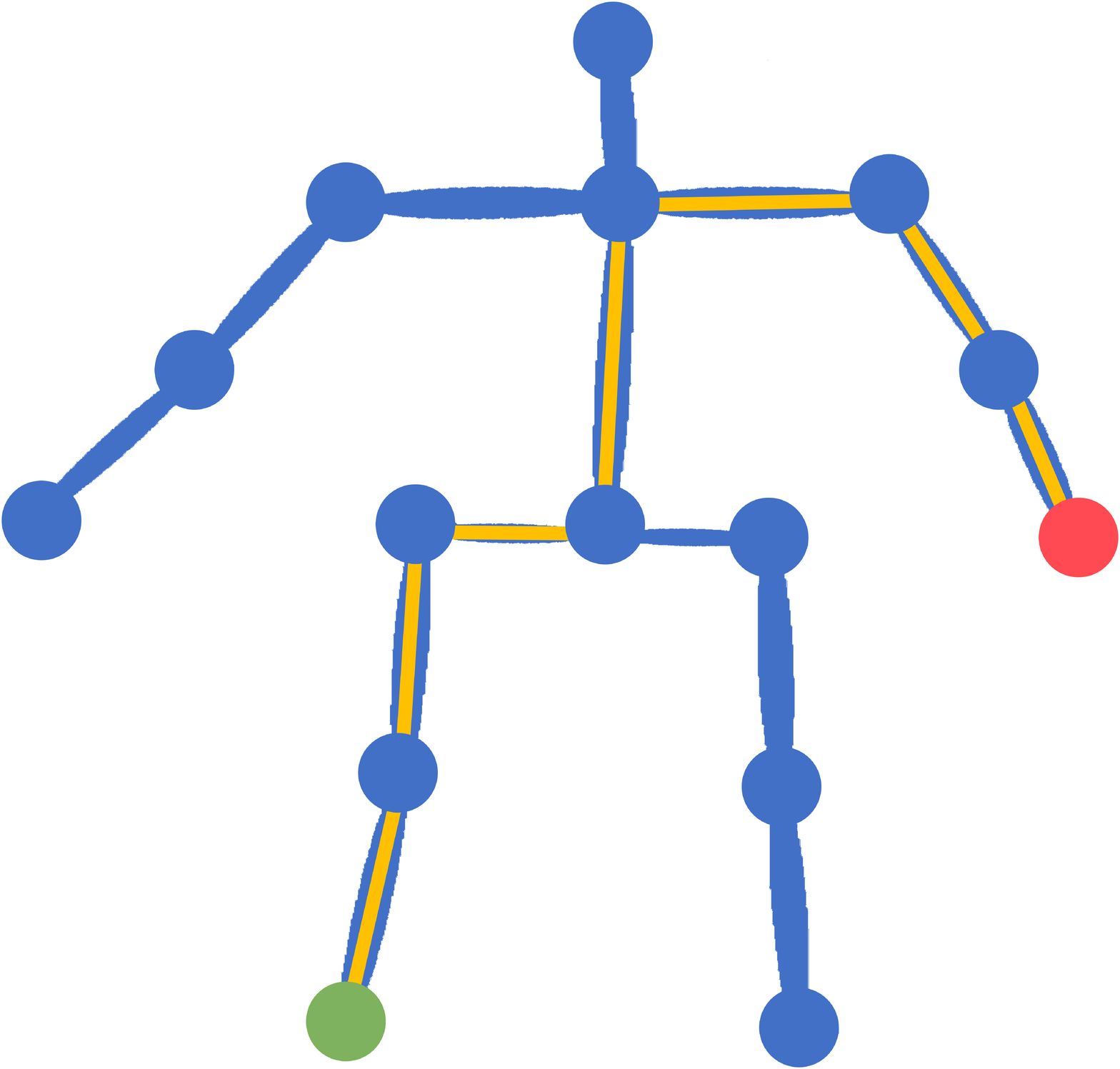}
\vspace{-5pt}
\caption{An illustration of the human joints. The red vertex denotes right wrist, the green vertex denotes left ankle, and the yellow lines denote the smallest hop between them.}
\label{fig:hop}
\vspace{-10pt}
\end{figure}

In current GCN-based works~\cite{liu2020disentangling}, to learn long-range dependencies, multi-scale graph convolution with multiple adjacency matrices are used to aggregate information of distant neighbors. The graph convolution can be formulated as
\begin{equation}\label{eq:1}
\mathbf{f}_\mathrm{out} = \sigma \left(\sum_{k}\mathbf{\Lambda}_k^{-\frac{1}{2}} \mathbf{A}_k \mathbf{\Lambda}_k^{-\frac{1}{2}}\mathbf{f}_\mathrm{in}\mathbf{W}_k \right), 
\end{equation}
where $\mathbf{f}_\mathrm{in} \in \mathbb{R}^{N \times C_1}$ and $\mathbf{f}_\mathrm{out} \in \mathbb{R}^{N \times C_2}$ denote the input and output feature maps on each frame, respectively. Here $C_1/C_2$ and $N$ denote the number of input/output feature channels and joints, respectively. 
Besides, $\mathbf{A}_k \in \{0,1\}^{N \times N}$ denotes the $k$-order adjacency matrix representing intra-frame connections, which is defined as
\begin{equation}\label{eq:4}
\mathbf{A}_k^{i,j}=\left\{
\begin{array}{l}
1, \quad \mathrm{if} \quad i=j \\
1, \quad  \mathrm{if} \quad d\left(v_i,v_j\right)=k \\
0, \quad \mathrm{otherwise}
\end{array}
\right.,
\end{equation}
where $d\left(v_i,v_j\right)$ denotes the smallest hop between nodes $v_i$ and $v_j$. $\mathbf{\Lambda}_k=\mathrm{diag}(\sum_{j}\mathbf{A}_k^{ij})  \in \mathbb{R}^{N \times N}$ is the degree matrix with the diagonal elements $\mathbf{\Lambda}_k^{ii} = \sum_{j}\mathbf{A}_k^{ij}$ and others as zeros, $\mathbf{W}_k \in \mathbb{R}^{C_1 \times C_2}$ is the learnable weight matrix and $\sigma \left(\cdot \right)$ is an activation function.

However, such method is suboptimal.
First, it need multiple matrices for modeling the joint dependence relation.
Take the right wrist (the red vertex in Figure~\ref{fig:hop}) for example, if all the other joints are taken into account, eight adjacency matrices are needed since the smallest hop between the right wrist and the farthest node, i.e., the left ankle (see the green vertex in Figure~\ref{fig:hop}), is seven. 
Second, we can see that all the adjacency matrixes include the self-connection (i.e., $i=j$), which makes the information of the right wrist itself will be aggregated eight times throughout the convolutional process, while other joints for only once. This may lead to a bias towards the local joint and the attenuation of other joint dependencies.
We argue that such factorized modeling is dispensable and the weight of all joints can be learned simultaneously. 

To address the problems, we propose a fully connected graph convolution operator to integrate multi-scale convolutions and balance the contributions from the joint itself and other joints. The operator no longer relies on the physical connections. Specifically, we define the adjacency matrix $\mathbf{A}$ as $\left\{\mathbf{A}^{ij}=1 \mid i,j=1,2,\dots,N \right\}$, that is, each pair of nodes of the skeleton graph is interconnected. We substitute $\mathbf{A}_k$ with $\mathbf{A}$ in Equation~\ref{eq:1} and obtain
\begin{equation}\label{eq:6}
\mathbf{f}_\mathrm{out} = \sigma \left( \mathbf{\Lambda} ^{-\frac{1}{2}} \mathbf{A} \mathbf{\Lambda} ^{-\frac{1}{2}}\mathbf{f}_\mathrm{in}\mathbf{W}\right), \quad \mathbf{\Lambda}=\mathrm{diag}(\sum_{j}\mathbf{A}^{ij}).
\end{equation}

Compared with multi-scale graph convolution where neighbors are divided into multiple scales based on the number of hops, the proposed fully connected convolution operator treats all nodes equally and automatically capture relationships between them.

\vspace{-3pt}
\subsection{Spatial-Temporal-Channel Attention (STC-Att)}
\label{sec:Attention}
\begin{figure}
\centering
\includegraphics[width=0.35\textwidth]{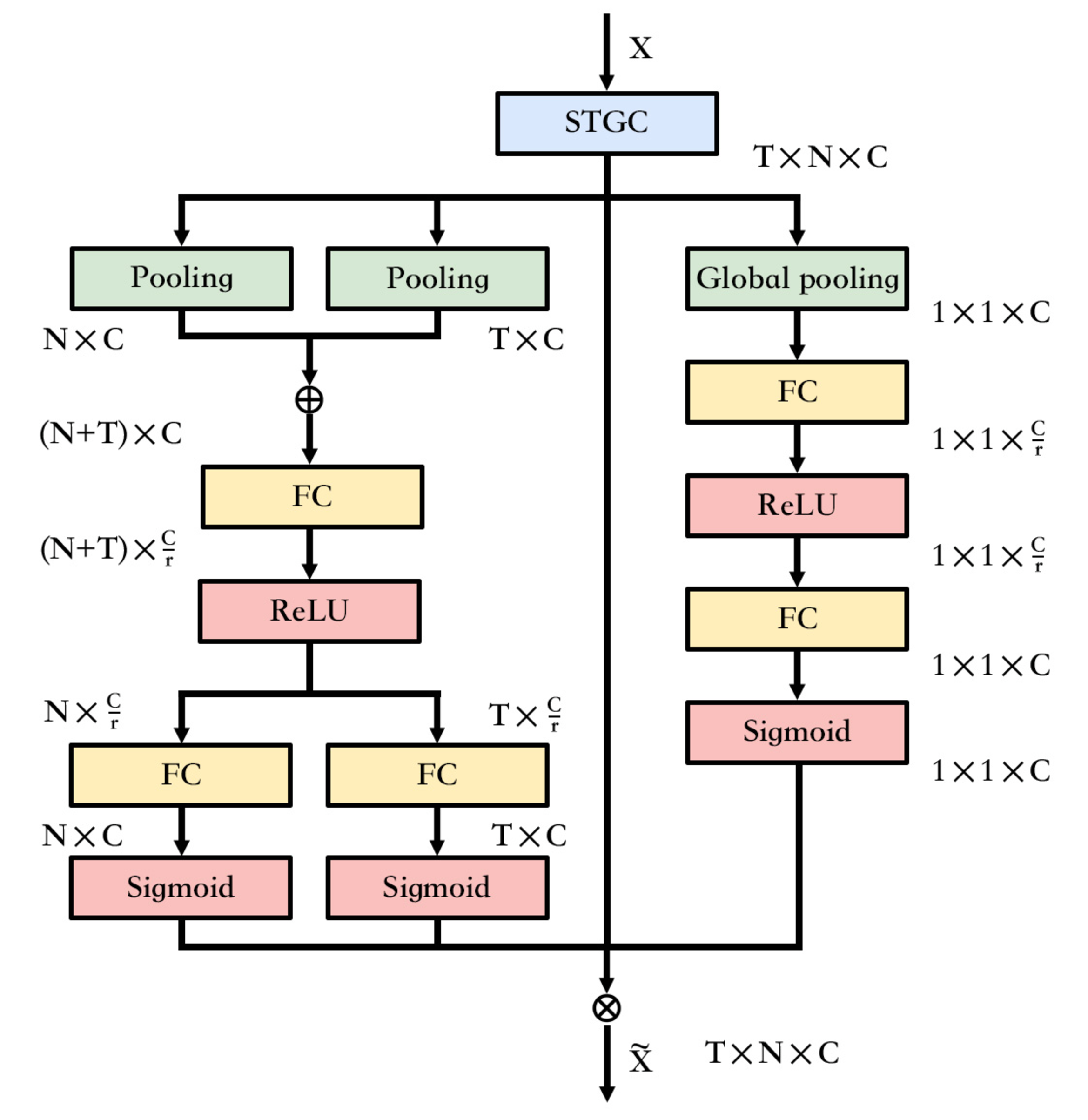}
\vspace{-10pt}
\caption{Schema of the proposed STC-Att module.}
\label{fig:att}
\vspace{-10pt}
\end{figure}

Previous works~\cite{wang2022multi,teepe2022towards} treat the information equally without distinguishment in spatial and temporal domains. 
Actually, not all frames and joints make the same contribution to gait recognition, \textit{i.e.}, the most discriminative information is commonly contained in some key joints from part of key frames.
Inspired by \cite{song2022constructing}, we propose a Spatial-Temporal-Channel Attention (STC-Att) module to dynamically adjust the weight of convolution kernel in spatial, temporal and channel domains. 

The concrete structure of the STC-Att module is illustrated in Figure~\ref{fig:att}. Channel-wise attention (the right branch) is performed according to the correlation between channels. 
Given a feature map $\textbf{f} \in {\mathbb{R}}^{T \times N\times C} $, the weight of each channel is learned in three steps. 1) The squeeze step, where global average pooling is performed to compact the feature map along the spatial and temporal dimension to a single value, that is, the feature map is converted into $\textbf{f}_\mathrm{c}\in {\mathbb{R}}^{1 \times 1 \times C}$. 2) The excitation step uses two FC layers with respective activation function to learn the nonlinear interaction among all channels, which finally generates the channel weight vector $\textbf{w}_\mathrm{c}\in {\mathbb{R}}^{1 \times 1 \times C}$ for the gait feature maps. 3) A scale layer is adopted to complete the recalibration of gait features through channel-wise product using the channel weight vector $\textbf{w}_\mathrm{c}$.

Spatial- and temporal-wise attentions (as shown in the left branch of Figure~\ref{fig:att}) are performed together to adaptively distinguish key joints and key frames from the whole skeleton sequence. Given a feature map $\textbf{f}\in {\mathbb{R}}^{T \times N\times C}$, we first average the spatial information along the spatial dimension into the feature map $\textbf{f}_\mathrm{t}\in {\mathbb{R}}^{T \times C}$, and the temporal information along the temporal dimension into the feature map $\textbf{f}_\mathrm{s}\in {\mathbb{R}}^{N\times C}$. 
Then, the pooled feature maps are concatenated together into feature map $\textbf{f}_\mathrm{st}\in {\mathbb{R}}^{ (V+T) \times C}$, followed by a FC layer and a ReLU activation function to reduce the feature dimension. 
After that, two independent FC layers followed by the activation functions are applied to learn the attention score vectors for temporal and spatial domains, respectively, which are denoted as $\textbf{w}_\mathrm{t}\in {\mathbb{R}}^{T \times C}$, $\textbf{w}_\mathrm{s}\in {\mathbb{R}}^{N\times C}$. Finally, the spatial/temporal attention vectors are multiplied on the input feature map, for spatial-temporal-adaptive feature refinement.

\vspace{-3pt}
\section{Experiments}
To evaluate our method, we conduct experiments on the public gait recognition dataset CASIA-B~\cite{yu2006framework} and OUMVLP~\cite{takemura2018multi}. 

\vspace{-3pt}
\subsection{Experimental Setting}
In model-based branch, the network is composed of 3 STGC blocks. The dimensions of output feature maps of each block are 96, 192, 384 in order.
The size of the raw skeleton data is $2\times 120\times 15$, which means that there are 120 frames for each sequence, 15 joints for each frame, and 2 channels (horizontal and vertical coordinate) for each joint. 
The batch size is 128 and the training epoch is 65. 
The initial learning rate is 0.1 and decays with a factor of 0.1 after the 45-th and 55-th epoch. 
In appearance-based branch, we follow the settings in~\cite{fan2020gaitpart}.

\vspace{-3pt}
\subsection{Comparisons with State-of-the-art Methods}
\begin{table*}
\centering
\caption{Comparison of gait recognition performance on CASIA-B.}
\label{tab:table4}
\resizebox{.88\width}{!}{
\begin{tabular}{clcccccccccccc}
\hline
\multicolumn{2}{c}{Gallery angle} & \multicolumn{12}{c}{$0^{\circ}-180^{\circ}$} \\ \hline
\multicolumn{2}{c}{Probe angle} & $0^{\circ}$ & $18^{\circ}$ & $36^{\circ}$ & \multicolumn{1}{c}{$54^{\circ}$} & \multicolumn{1}{c}{$72^{\circ}$} & \multicolumn{1}{c}{$90^{\circ}$} & $108^{\circ}$ & $126^{\circ}$ & $144^{\circ}$ & \multicolumn{1}{c}{$162^{\circ}$} & \multicolumn{1}{c}{$180^{\circ}$} & \multicolumn{1}{c}{Mean} \\ \hline
\multirow{5}{*}{NM}
& GaitGraph2~\cite{teepe2022towards} & 78.5 & 82.9 & 85.8 & 85.6 & 83.1 & 81.5 & 84.3 & 83.2 & 84.2 & 81.6 & 71.8 & 82.0 \\
 & FR-GCN~\cite{wang2022frame} & 90.6 & 91.6 & 93.5 & 92.3 & 91.9 & 93.0 & 92.1 & 91.6 & 93.3 & 89.3 & 86.0 & 91.4 \\
 & GaitGL~\cite{lin2021gait} & 96.0 & \textbf{98.3} & \textbf{99.0} & 97.9 & 96.9 & 95.4 & 97.0 & \textbf{98.9} & \textbf{99.3} & \textbf{98.8} & 94.0 & 97.4 \\
 & Lagrange~\cite{chai2022lagrange} & 95.2 & 97.8 & \textbf{99.0} & 98.0 & 96.9 & 94.6 & 96.9 & 98.8 & 98.9 & 98.0 & 91.5 & 96.9 \\
 & Ours & \textbf{97.0} & 97.9 & 98.4 & \textbf{98.3} & \textbf{97.2} & \textbf{97.3} & \textbf{98.2} & 98.4 & 98.3 & 98.1 & \textbf{96.0} & \textbf{97.7} \\ \hline
\multirow{5}{*}{BG} 
& GaitGraph2~\cite{teepe2022towards} & 69.9 & 75.9 & 78.1 & 79.3 & 71.4 & 71.7 & 74.3 & 76.2 & 73.2 & 73.4 & 61.7 & 73.2 \\
 & FR-GCN~\cite{wang2022frame} & 77.9 & 85.2 & 84.0 & 81.2 & 82.5 & 78.9 & 81.3 & 79.5 & 80.2 & 77.8 & 71.0 & 80.0 \\
 & GaitGL~\cite{lin2021gait} & \textbf{92.6} & \textbf{96.6} & \textbf{96.8} & \textbf{95.5} & 93.5 & 89.3 & 92.2 & \textbf{96.5} & \textbf{98.2} & 96.9 & \textbf{91.5} & \textbf{94.5} \\
 & Lagrange~\cite{chai2022lagrange} & 89.9 & 94.5 & 95.9 & 94.6 & 93.9 & 88.0 & 91.1 & 96.3 & 98.1 & \textbf{97.3} & 88.9 & 93.5 \\
 & Ours & 91.9 & 94.6 & 96.4 & 94.3 & \textbf{94.4} & \textbf{91.6} & \textbf{94.1} & 95.4 & 95.5 & 93.9 & 89.5 & 93.8 \\ \hline
\multirow{5}{*}{CL}
& GaitGraph2~\cite{teepe2022towards} & 57.1 & 61.1 & 68.9 & 66.0 & 67.8 & 65.4 & 68.1 & 67.2 & 63.7 & 63.6 & 50.4 & 63.6 \\
 & FR-GCN~\cite{wang2022frame} & 74.0 & 74.3 & 76.0 & 78.8 & 80.4 & 79.3 & 78.0 & 79.5 & 74.8 & 70.5 & 67.0 & 75.7 \\
 & GaitGL~\cite{lin2021gait} & 76.6 & 90.0 & 90.3 & 87.1 & 84.5 & 79.0 & 84.1 & 87.0 & 87.3 & 84.4 & 69.5 & 83.6 \\
 & Lagrange~\cite{chai2022lagrange} & 81.6 & 91.0 & 94.8 & 92.2 & 85.5 & 82.1 & 86.0 & 89.8 & 90.6 & 86.0 & 73.5 & 86.6 \\
 & Ours & \textbf{87.4} & \textbf{96.0} & \textbf{97.0} & \textbf{94.6} & \textbf{94.0} & \textbf{90.1} & \textbf{91.5} & \textbf{94.1} & \textbf{93.8} & \textbf{92.6} & \textbf{88.5} & \textbf{92.7} \\ \hline
\end{tabular}
}\vspace{-15pt}
\end{table*}

\begin{table*}
\centering
\caption{Comparison of gait recognition performance on OUMVLP.}
\label{tab:oumvlp}
\resizebox{.95\width}{!}{
\begin{tabular}{lccccccccccccccc}
\hline
\multicolumn{1}{c}{Probe angle} & $0^{\circ}$ & $15^{\circ}$ & $30^{\circ}$ & \multicolumn{1}{c}{$45^{\circ}$} & \multicolumn{1}{c}{$60^{\circ}$} & \multicolumn{1}{c}{$75^{\circ}$} & $90^{\circ}$ & $180^{\circ}$ & $195^{\circ}$ & \multicolumn{1}{c}{$210^{\circ}$} & \multicolumn{1}{c}{$225^{\circ}$} & \multicolumn{1}{c}{$240^{\circ}$} & \multicolumn{1}{c}{$255^{\circ}$} & \multicolumn{1}{c}{$270^{\circ}$} & \multicolumn{1}{c}{Mean} \\ \hline
GaitGraph2~\cite{teepe2022towards} & 54.3 & 68.4 & 76.1 & 76.8 & 71.5 & 75.0 & 70.1 & 52.2 & 60.6 & 57.8 & 73.2 & 67.8 & 70.8 & 65.3 & 67.1 \\
FR-GCN~\cite{wang2022frame} & 48.3 & 53.5 & 56.8 & 58.9 & 58.3 & 55.2 & 50.6 & 36.6 & 49.0 & 45.5 & 60.6 & 60.4 & 57.4 & 53.6 & 53.2 \\
GaitGL~\cite{lin2021gait} & 84.9 & 90.2 & 91.1 & \textbf{91.5} & 91.1 & 90.8 & \textbf{90.3} & 88.5 & 88.6 & 90.3 & \textbf{90.4} & 89.6 & 89.5 & 88.8 & 89.7 \\
Lagrange~\cite{chai2022lagrange} & 84.5 & 89.8 & 91.0 & 91.2 & 90.7 & 90.5 & 90.2 & 88.5 & 87.9 & 89.9 & 90.0 & 89.2 & 89.2 & 88.7 & 89.4 \\
Ours & \textbf{91.3} & \textbf{92.4} & \textbf{91.2} & 89.9 & \textbf{92.1} & \textbf{90.9} & 90.2 & \textbf{90.0} & \textbf{92.1} & \textbf{90.3} & 89.5 & \textbf{92.5} & \textbf{90.7} & \textbf{90.5} & \textbf{91.0} \\ \hline
\end{tabular}
}
\vspace{-22pt}
\end{table*}

We compare the proposed two-branch network to state-of-the-art gait recognition methods with the same experimental settings comprehensively, including GaitGraph2~\cite{teepe2022towards}, FR-GCN~\cite{wang2022frame}, GaitGL~\cite{lin2021gait} and Lagrange~\cite{chai2022lagrange}. Among that, the first two belong to model-based approaches and the last two belong to appearance-based approaches. Note that, the accuracies of the compared approaches are directly cited from their original papers.
The detailed results are listed in Table~\ref{tab:table4} and \ref{tab:oumvlp}. 
We can see that our model presents the superior performance under various conditions. 
Especially in the CL (Clothing) condition on CASIA-B, the accuracy of ours increases by 9.1\% compared with GaitGL~\cite{lin2021gait}, which can indicate that the proposed network have great effectiveness in handling with occlusion and abstracting discriminative gait descriptors.

\vspace{-3pt}
\subsection{Ablation Study}
\label{sec:Ablation}

\vspace{-4pt}
\textbf{Effectiveness of components in GCN.}
We validate the effectiveness of each proposed component in GCN-based branch in Table~\ref{tab:table2}.
From the comparison between the first and second rows, we can clearly seen that the model with fully connected graph convolution gains higher accuracies for all conditions, which verifies the availability of this operator.
In addition, it can be seen that with the integration of our attention mechanism (from the second row to the third row), the recognition accuracies of the network have been improved in all conditions, which demonstrates the superiority of STC-Att for capturing the discriminative gait representations.
Note that all the results in Table~\ref{tab:table2} are generated using only skeleton features.

\begin{table}
\centering
\caption{Ablation studies conducted on CASIA-B.}
\label{tab:table2}
\resizebox{.95\width}{!}{
\begin{tabular}{lccc}
\hline
 & \multicolumn{3}{c}{Accuracy(\%)}  \\ \cline{2-4} 
 & NM  & BG  & CL   \\ \hline
GCN-Baseline & 89.8 & 78.5 & 78.0 \\
+ Fully-connected & 91.3 & 79.4 & 78.5 \\
+ STC-Att (Ours)  & \textbf{91.8} & \textbf{79.8} & \textbf{79.4} \\ \hline
\end{tabular}}
\vspace{-15pt}
\end{table} 

\textbf{Discussion of two-branch fusion.}
The accuracies using different networks are shown in Table~\ref{tab:table3}. 
The first two lines show the averaged accuracies of model- and appearance-based branch, respectively. 
As mentioned in Section~\ref{sec:Model}, to combine the model- and appearance-based features, a ratio $\lambda$ is needed to balance the disparity between them.
The following five lines show the performance of the proposed two-branch network using different values of $\lambda$.
We can see that the proposed two-branch network consistently gets a better performance with a large margin than using only one of them.

In order to determine the most suitable ratio of the two-branch network, we compare different values of $\lambda$ from 300 to 500 with 50 as interval. 
We observe that as the value of $\lambda$ increases, the accuracy rate under all conditions shows a trend of first increasing and then decreasing. NM and BG condition get the best results with setting $\lambda =450$ and CL condition gets the best results with setting $\lambda =300$. In addition, we calculate the mean value of accuracies under the three conditions (shown in the last column). Considering that the model should perform well in various states, we choose $\lambda =400$ with the highest mean value as the final ratio.

\begin{table}
\centering
\caption{Averaged accuracies with different values of $\lambda$.}
\label{tab:table3}
\resizebox{.9\width}{!}{
\begin{tabular}{lcccc}
\hline
\multicolumn{1}{l}{}  & \multicolumn{4}{c}{Accuracy(\%)} \\ \cline{2-5} 
\multicolumn{1}{l}{}    & NM & BG & CL & Mean \\ \hline
model-based branch      & 91.8 & 79.8 & 79.4 & 83.67  \\
appearance-based branch & 91.5 & 81.7 & 68.6 & 80.60 \\ \hline
$\lambda =300$ & 97.0 & 92.8 & \textbf{93.4} & 94.38 \\
$\lambda =350$  & 97.5 & 93.4  & 93.1 & 94.65   \\
$\lambda =400$ (Ours)  & 97.7 & 93.8 & 92.7 & \textbf{94.74} \\
$\lambda =450$  & \textbf{97.7} & \textbf{93.9} & 91.9  & 94.51  \\
$\lambda =500$  & 97.6 & 93.5 & 90.8  & 93.98 \\ \hline
\end{tabular}}
\vspace{-15pt}
\end{table} 

\vspace{-3pt}
\section{Conclusion}
\vspace{-3pt}
In this paper, we have proposed a simple yet effective two-branch network, containing a CNN-based branch extracting gait features from silhouettes and a GCN-based branch extracting gait features from skeletons. We also provided a fully connected graph convolution operator and an attention module for efficient and effective gait feature representation. 
Extensive experiments show that our method outperforms the state-of-the-art methods under all conditions, which show its great potential in dealing with the challenges caused by the viewpoint, carrying and clothing condition variations.


%



\bibliographystyle{IEEEbib}
\bibliography{strings,refs}

\begin{thebibliography}{10}

\bibitem{huang2021context}
Xiaohu Huang, Duowang Zhu, Hao Wang, Xinggang Wang, Bo~Yang, Botao He, Wenyu
  Liu, and Bin Feng,
\newblock ``Context-sensitive temporal feature learning for gait recognition,''
\newblock in {\em ICCV}, 2021, pp. 12909--12918.

\bibitem{xu2020cross}
Chi Xu, Yasushi Makihara, Xiang Li, Yasushi Yagi, and Jianfeng Lu,
\newblock ``Cross-view gait recognition using pairwise spatial transformer
  networks,''
\newblock {\em IEEE TCSVT}, vol. 31, no. 1, pp. 260--274, 2020.

\bibitem{li2019joint}
Xiang Li, Yasushi Makihara, Chi Xu, Yasushi Yagi, and Mingwu Ren,
\newblock ``Joint intensity transformer network for gait recognition robust
  against clothing and carrying status,''
\newblock {\em IEEE TIFS}, vol. 14, no. 12, pp. 3102--3115, 2019.

\bibitem{wu2016comprehensive}
Zifeng Wu, Yongzhen Huang, Liang Wang, Xiaogang Wang, and Tieniu Tan,
\newblock ``A comprehensive study on cross-view gait based human identification
  with deep cnns,''
\newblock {\em IEEE TPAMI}, vol. 39, no. 2, pp. 209--226, 2016.

\bibitem{zhang2019learning}
Kaihao Zhang, Wenhan Luo, Lin Ma, Wei Liu, and Hongdong Li,
\newblock ``Learning joint gait representation via quintuplet loss
  minimization,''
\newblock in {\em CVPR}, 2019, pp. 4700--4709.

\bibitem{chao2019gaitset}
Hanqing Chao, Yiwei He, Junping Zhang, and Jianfeng Feng,
\newblock ``Gaitset: Regarding gait as a set for cross-view gait recognition,''
\newblock in {\em AAAI}, 2019, pp. 8126--8133.

\bibitem{su2020deep}
Jingran Su, Yang Zhao, and Xuelong Li,
\newblock ``Deep metric learning based on center-ranked loss for gait
  recognition,''
\newblock in {\em ICASSP}, 2020, pp. 4077--4081.

\bibitem{lin2021gait}
Beibei Lin, Shunli Zhang, and Xin Yu,
\newblock ``Gait recognition via effective global-local feature representation
  and local temporal aggregation,''
\newblock in {\em ICCV}, 2021, pp. 14648--14656.

\bibitem{chai2022lagrange}
Tianrui Chai, Annan Li, Shaoxiong Zhang, Zilong Li, and Yunhong Wang,
\newblock ``Lagrange motion analysis and view embeddings for improved gait
  recognition,''
\newblock in {\em CVPR}, 2022, pp. 20249--20258.

\bibitem{cao2017realtime}
Zhe Cao, Tomas Simon, Shih-En Wei, and Yaser Sheikh,
\newblock ``Realtime multi-person 2d pose estimation using part affinity
  fields,''
\newblock in {\em CVPR}, 2017, pp. 7291--7299.

\bibitem{an2018improving}
Weizhi An, Rijun Liao, Shiqi Yu, Yongzhen Huang, and Pong~C Yuen,
\newblock ``Improving gait recognition with 3d pose estimation,''
\newblock in {\em Chinese Conference on Biometric Recognition}, 2018, pp.
  137--147.

\bibitem{liao2020model}
Rijun Liao, Shiqi Yu, Weizhi An, and Yongzhen Huang,
\newblock ``A model-based gait recognition method with body pose and human
  prior knowledge,''
\newblock {\em PR}, vol. 98, pp. 107069, 2020.

\bibitem{teepe2022towards}
Torben Teepe, Johannes Gilg, Fabian Herzog, Stefan H{\"o}rmann, and Gerhard
  Rigoll,
\newblock ``Towards a deeper understanding of skeleton-based gait
  recognition,''
\newblock in {\em CVPRW}, 2022, pp. 1569--1577.

\bibitem{wang2022frame}
Likai Wang, Jinyan Chen, and Yuxin Liu,
\newblock ``Frame-level refinement networks for skeleton-based gait
  recognition,''
\newblock {\em CVIU}, vol. 222, pp. 103500, 2022.

\bibitem{cai2021hybrid}
Ning Cai, Shiling Feng, Qing Gui, Lei Zhao, Huadong Pan, Jun Yin, and Bin Lin,
\newblock ``Hybrid silhouette-skeleton body representation for gait
  recognition,''
\newblock in {\em IHMSC}, 2021, pp. 216--220.

\bibitem{wang2022multi}
Likai Wang, Jinyan Chen, Zhenghang Chen, Yuxin Liu, and Haolin Yang,
\newblock ``Multi-stream part-fused graph convolutional networks for
  skeleton-based gait recognition,''
\newblock {\em Connection Science}, vol. 34, no. 1, pp. 652--669, 2022.

\bibitem{li2019attentive}
Shuangqun Li, Wu~Liu, and Huadong Ma,
\newblock ``Attentive spatial--temporal summary networks for feature learning
  in irregular gait recognition,''
\newblock {\em IEEE TMM}, vol. 21, no. 9, pp. 2361--2375, 2019.

\bibitem{dou2022metagait}
Huanzhang Dou, Pengyi Zhang, Wei Su, Yunlong Yu, and Xi~Li,
\newblock ``Metagait: Learning to learn an omni sample adaptive representation
  for gait recognition,''
\newblock in {\em ECCV}, 2022, pp. 357--374.

\bibitem{liu2020disentangling}
Ziyu Liu, Hongwen Zhang, Zhenghao Chen, Zhiyong Wang, and Wanli Ouyang,
\newblock ``Disentangling and unifying graph convolutions for skeleton-based
  action recognition,''
\newblock in {\em CVPR}, 2020, pp. 143--152.

\bibitem{fan2020gaitpart}
Chao Fan, Yunjie Peng, Chunshui Cao, Xu~Liu, Saihui Hou, Jiannan Chi, Yongzhen
  Huang, Qing Li, and Zhiqiang He,
\newblock ``Gaitpart: Temporal part-based model for gait recognition,''
\newblock in {\em CVPR}, 2020, pp. 14225--14233.

\bibitem{song2022constructing}
Yi-Fan Song, Zhang Zhang, Caifeng Shan, and Liang Wang,
\newblock ``Constructing stronger and faster baselines for skeleton-based
  action recognition,''
\newblock {\em IEEE TPAMI}, vol. 45, no. 2, pp. 1474--1488, 2022.

\bibitem{yu2006framework}
Shiqi Yu, Daoliang Tan, and Tieniu Tan,
\newblock ``A framework for evaluating the effect of view angle, clothing and
  carrying condition on gait recognition,''
\newblock in {\em ICPR}, 2006, pp. 441--444.

\bibitem{takemura2018multi}
Noriko Takemura, Yasushi Makihara, Daigo Muramatsu, Tomio Echigo, and Yasushi
  Yagi,
\newblock ``Multi-view large population gait dataset and its performance
  evaluation for cross-view gait recognition,''
\newblock {\em IPSJ transactions on Computer Vision and Applications}, vol. 10,
  pp. 1--14, 2018.

\end{thebibliography}

\end{document}